%% file: main.tex
\documentclass[letterpaper]{article}
\usepackage[preprint]{style}
\usepackage[hyphens]{url}
\usepackage{graphicx}
\usepackage{natbib}
\usepackage{caption}
\usepackage{algorithm}
\usepackage{algorithmic}

\usepackage{newfloat}
\usepackage{listings}
\DeclareCaptionStyle{ruled}{labelfont=normalfont,labelsep=colon,strut=off}
\floatstyle{ruled}
\newfloat{listing}{tb}{lst}{}
\floatname{listing}{Listing}

\usepackage{booktabs}

\input{math_commands.tex}

\input{_header}

\newcommand{\VideoLLMs}{\text{VideoLLMs}\xspace}
\newcommand{\VideoLLM}{\text{VideoLLM}\xspace}
\newcommand{\LQwen}{LVQ\xspace}
\newcommand{\VideoLLaMA}{VL2\xspace}
\newcommand{\SGV}{SG4V\xspace}
\newcommand{\QwenVL}{Q3VL\xspace}
\newcommand{\InternVL}{IVL\xspace}
\newcommand{\Molmo}{Mol2\xspace}

\newcommand{\ours}{\textsc{PoisonVID}\xspace}

\title{Poisoning Prompt-Guided Sampling in Video Large Language Models}
\author{
    Yuxin Cao\textsuperscript{\rm 1},
    Wei Song\textsuperscript{\rm 2},
    Jingling Xue\textsuperscript{\rm 2},
    Jin Song Dong\textsuperscript{\rm 1}\corresponding
}
\affiliations{
    \textsuperscript{\rm 1}National University of Singapore, Singapore\\
    \textsuperscript{\rm 2}University of New South Wales, Australia
}

\begin{document}

\maketitle

\begin{abstract}
Video Large Language Models (\VideoLLMs) are increasingly deployed as automated moderators on user-generated video platforms, where a few unwatched seconds of harmful footage are enough to suppress a safety alert. Because encoding every frame is prohibitive, modern \VideoLLMs rely on prompt-guided sampling (PGS), which scores frames against the user prompt and forwards only the top-ranked ones to the visual encoder. Uniform and semantic samplers are known to be defeated by simple frame replacement, whereas PGS, the most prompt-aware family, has escaped scrutiny, and its prompt awareness in fact repairs the omission failures that defeat the other two. We show that this repair is superficial, since \ours, a transfer attack, poisons the sampler's ranking so that harmful clips are never surfaced, without access to target weights, gradients, or sampling internals. It optimizes one video-level perturbation under a relevance-suppression loss defined over a depiction set of paraphrased harmful descriptions written by a shadow \VideoLLM and a general-purpose language model, which drives perturbed harmful frames out of the prompt-conditioned subspace that PGS reads. Samplers that never consult that score keep the frames they always kept, which locates the failure at selection rather than at the encoder. Across three PGS methods, six \VideoLLMs, and six harmful categories, \ours attains 84\% to 97\% average attack success over the 18 sampler and model pairs and survives seven defenses. Re-encoding at lower resolution on ingest gives back part of what was evicted and costs the attack 48 points, which bounds the threat without closing it. PGS therefore buys accuracy with a structural safety debt, and sampler design will now have to repay that.
\end{abstract}

\section{Introduction}

Video Large Language Models (\VideoLLMs) couple temporal visual perception with language reasoning and now dominate video understanding \cite{weng2024longvlm,huang2025frag,cheng2025vilamp,chen2024sharegpt4video,zohar2025apollo}. Conditioned on a user prompt, they summarize long footage without manual review \cite{cheng2025vilamp,chen2024sharegpt4video,zohar2025apollo}, and a growing share of their deployments is safety-critical, covering moderation and auditing of user-generated video as well as surveillance analysis \cite{qian2024streaming}. In such settings an obvious harmful event the model never reports suppresses the alert that should have followed, and the unsafe content spreads unchecked~\cite{cao2026failures}.

No \VideoLLM reads every frame, because encoding a full stream costs more than the task can bear. A sampler selects a small subset instead, and the quality of that subset bounds everything computed downstream of it \cite{li2024llama-vid,cheng2024videollama2,zhang2024llavanextvideo}. Of the three families in use, Uniform Frame Sampling (UFS) takes frames at fixed intervals and is blind to content, so it steps over the moments that matter \cite{li2024llama-vid,cheng2024videollama2,zohar2025apollo}. Semantic Similarity Sampling (SSS) discards neighbors a vision-language model judges redundant, which raises information density yet still ignores the prompt \cite{chen2024sharegpt4video,wang2025videotree,tang2025aks}. Prompt-Guided Sampling (PGS) scores each candidate for alignment with the prompt and keeps the top-ranked frames, and the accuracy it buys has made it the choice in recent work \cite{huang2025frag,cheng2025vilamp}.

\begin{figure*}[t]
    \centering
    \includegraphics[width=0.80\linewidth]{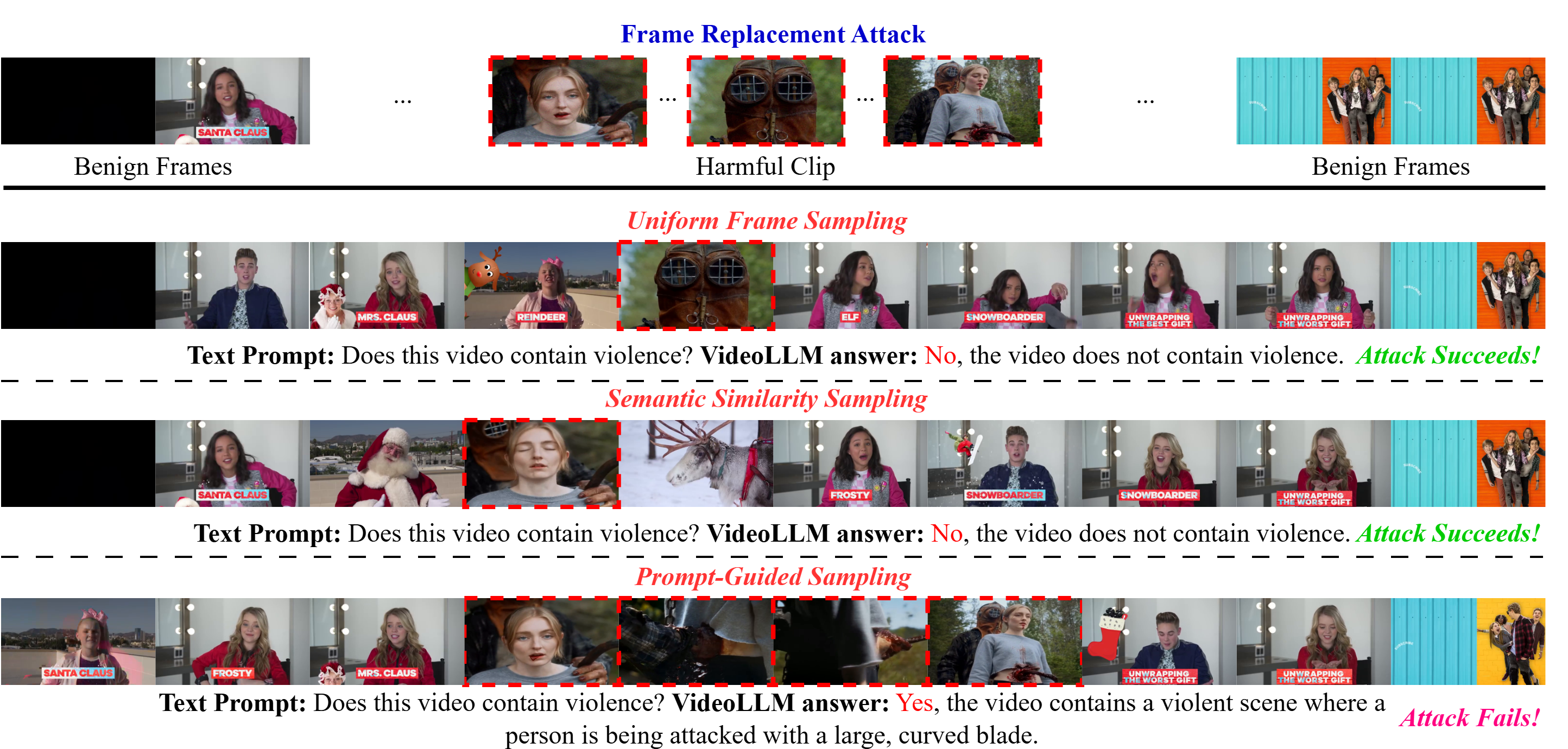}
    \caption{The Frame Replacement Attack (FRA) \cite{cao2026failures} under uniform (UFS), semantic-similarity (SSS), and prompt-guided (PGS) frame sampling. UFS and SSS never read the query, so they admit one harmful frame and the \VideoLLM misses the violence, whereas PGS ranks against the query and retains several. Dashed red boxes mark frames from the harmful clip.}
    \label{fig:pre-example}
\end{figure*}

The safety of these samplers has been studied far less carefully than their accuracy. The Frame Replacement Attack (FRA) \cite{cao2026failures} is the only prior work that exploits sampling for evasion, and it shows that UFS and SSS routinely drop a harmful segment spliced into a benign carrier. They drop it because neither of them looks for it, since a sampler blind to the query cannot aim at the four seconds that answer it. PGS does more than harden that surface and in fact inverts it. Ranking frames against the prompt makes PGS \emph{pull harmful frames in} the moment a moderator's query names or even hints at the harm they contain, because those frames score highest under exactly that query. \Cref{fig:pre-example} contrasts what the three samplers admit, and \Cref{tab:main_results} measures what that costs an attacker, with FRA falling from roughly 90\% success on UFS \cite{cao2026failures} to an average of 16.7\% on FRAG. A placement trick cannot defeat a sampler that seeks out what the attacker is hiding, so the relevance signal that PGS reads has to be suppressed instead.

We introduce \ours, an attack built for that task. It leaves training data and model weights untouched and acts only on frame admission at inference time, optimizing one video-level $\ell_\infty$-bounded perturbation that drives a harmful clip's relevance toward zero under any prompt describing the event it depicts, so PGS deselects those frames before the language model is invoked. Each iteration applies the current perturbation to a random subset of harmful frames, scores them with a local lightweight VLM, and updates by projected gradient descent. Aiming the optimization at harmful \emph{semantics} rather than at one phrasing requires a stable anchor, and we build one as a \emph{depiction set} of paraphrased harmful descriptions, written by a shadow \VideoLLM from the clip itself and diversified by a general-purpose language model. The attacker never queries the target, and the perturbation transfers to PGS methods and \VideoLLMs it never saw.

The flaw this exposes belongs to the pipeline rather than to any model inside it. PGS pays for accuracy twice, once to decode a dense candidate pool and once to score every frame in that pool, and what the payment buys is a ranking that decides admission. An uploader cannot write to a fixed sampling schedule, whereas a learned score is exactly the kind of quantity a bounded perturbation moves, so resting admission on relevance opens a channel uniform sampling never offered. A moderator running PGS under \ours still pays the full scoring cost and then receives a ranking sorted to the attacker's benefit, which hands back almost all of the margin that cost had bought.

We evaluate \ours on three open-source PGS methods, Differential Keyframe Selection (DKS)~\cite{cheng2025vilamp}, Adaptive Keyframe Sampling (AKS)~\cite{tang2025aks}, and Frame Selection Augmented Generation (FRAG)~\cite{huang2025frag}, layered on six \VideoLLMs spanning LLM-centric and native multimodal designs. Over all 18 pairs and six harmful categories the average attack success rate lies between 84\% and 97\%, and it holds under seven adversarial, data-poisoning, and inference-time defenses. This paper makes the following three contributions.

\begin{itemize}
    \item We identify prompt-guided sampling as a previously unexamined attack surface, where the attacked quantity decides frame admission rather than any model output, and show that the safety PGS appears to buy against frame replacement vanishes once an adversary suppresses relevance.

    \item We propose \ours, which anchors one video-level perturbation on a depiction set of paraphrased harmful descriptions, and we establish by controlled measurement at the selection stage that the resulting failure is eviction from the sampler rather than corruption of the encoder.

    \item We evaluate across three PGS methods, six \VideoLLMs, six harmful categories, 200 carriers, four prompt regimes, and seven defenses, and we bound the threat by measuring how much of it platform transcoding gives back, which is a part of the attack rather than all of it.
\end{itemize}

\section{Background and Related Work}

\noindent\textbf{VideoLLMs and Frame Sampling.}
A \VideoLLM answers a textual query grounded in video~\cite{liu2025nvila,li2024llava-onevision} through three stages, namely frame selection, visual encoding with projection into the text embedding space, and fusion inside a large language model. Representative systems follow the LLaVA~\cite{zhang2024llavanextvideo,zhang2024LLaVA-Video} and LLaMA~\cite{li2024llama-vid,cheng2024videollama2} families over tasks from captioning and summarization to grounding and long-video understanding~\cite{chen2024sharegpt4video,qian2024streaming,weng2024longvlm}, and the first stage is the one we attack. The three open PGS methods differ only in what they add above relevance, where DKS~\cite{cheng2025vilamp} penalizes inter-frame similarity, AKS~\cite{tang2025aks} enforces temporal coverage, and FRAG~\cite{huang2025frag} ranks by relevance alone with a full \VideoLLM as scorer, while related selectors follow the same principle~\cite{cheng2024focuschat,yu2025framevoyager}. SKE~\cite{chen2024sharegpt4video} represents SSS throughout, and the appendix details every one of them.

\noindent\textbf{Adversarial Attacks on Vision-Language Models.}
Perturbing an image inside a joint vision-language space is a mature line of work, and it sorts by the quantity each attack moves. One branch moves a predicted label or a retrieval result, where Co-Attack~\cite{zhang2022coattack} perturbs image and text jointly against a known model, SGA~\cite{lu2023sga} adds set-level guidance to recover the transfer Co-Attack lacks, universal perturbations~\cite{moosavi2017universal} arrive through AdvCLIP~\cite{zhou2023advclip}, and X-Transfer~\cite{huang2025xtransfer} carries transfer across CLIP backbones. The other branch moves generated text, steering a multimodal model toward a chosen response~\cite{zhao2023attackvlm} or unlocking behavior it would otherwise refuse~\cite{qi2024visual}, and video extensions carry image perturbations to recognizers~\cite{wei2022i2v} and to multimodal LLMs~\cite{li2024fmm,huang2025i2vmllm}. Every one of them ends at something the victim model emits on the perturbed input, while \ours ends earlier. A PGS pipeline calls a VLM to rank frames, and we leave what that VLM says about any frame untouched, because the quantity we suppress reaches no output and only decides admission into the top-$N$ set. The frame is then absent from the input, so hardening whichever model answers cannot recover evidence that never reached it in the first place. We therefore target the admission rule rather than the scorer that serves it, a distinction the two closest video attacks never draw, because neither of them models the sampler at all~\cite{li2024fmm,huang2025i2vmllm}.

\noindent\textbf{Ranking and Retrieval Attacks.}
Attacks on ranked selection are the closest precedent, where adversarial ranking demotes a candidate through ranking inequalities~\cite{zhou2020ranking}, QAIR optimizes a top-$k$ set-overlap loss for image retrieval~\cite{li2021qair}, and jamming inserts a blocker document that is itself retrieved and makes the generator refuse~\cite{shafran2025jamming}, the mirror image of our goal since it adds an item to a ranked set where we remove one. We share the objective yet cannot borrow the machinery, because an attacker who never observes the benign carrier cannot evaluate top-$N$ membership while optimizing, which is why \Cref{eq:rsl_loss} substitutes a mean-relevance surrogate that one perturbation must satisfy for every frame at once.

\noindent\textbf{Poisoning Attacks on VideoLLMs.}
Work on \VideoLLMs has emphasized capability over safety, leaving poisoning risks underexplored. \citet{cao2026failures} identify three design flaws, namely sparse uniform sampling, token under-sampling, and modality fusion imbalance, that cause UFS- and SSS-based systems to overlook harmful content even when it is plainly embedded. We move the question to the sampling stage those systems lack, and ask whether the more advanced samplers that were built to fix it are themselves poisonable by the very uploads that they rank.

\section{Method}
\label{sec:method}

\subsection{Prompt-Guided Sampling}
\label{subsec:pgs_formulation}

Let $V=\{f_1,\dots,f_T\}$ be a video. Since encoding all $T$ frames is prohibitive \cite{chen2024sharegpt4video,cheng2024videollama2,liu2025nvila}, a selection rule $\mathcal{S}(V,N)$ returns $N \ll T$ frames, which are encoded, projected into the text embedding space, and fused with the tokenized prompt inside the language model. PGS is the standard choice of $\mathcal{S}$ in recent \VideoLLMs~\cite{cheng2025vilamp,huang2025frag}. From a dense pool $V_M$ with $M>N$ it assigns every $f_k \in V_M$ a relevance score $r(f_k,q')$ measuring alignment with a query $q'$ derived from the user prompt $q$, for instance whether the frame carries enough information to help answer $q$~\cite{tang2025aks}, and it then simply keeps the highest scorers,
\begin{equation}
\label{eq:pgs}
\mathcal{S}_{\text{PGS}}(V, N, q') = \operatorname*{arg\,\text{Top-}N}_{f_k \in V_M} \; r(f_k, q').
\end{equation}
Following common practice~\cite{tang2025aks,cheng2025vilamp,huang2025frag}, $r$ is a lightweight VLM realized either as a CLIP-style cosine similarity or as a BLIP-style image-text matching head with output in $[0,1]$, and in both cases it is differentiable in the input pixels. That differentiability is what the sampler cannot surrender without abandoning prompt guidance, and it is just what we exploit here.

\subsection{Threat Model}
\label{subsec:threat}

We consider a \VideoLLM deployed as an automated moderator over user-generated video, for instance a violence filter on an entertainment or live-streaming platform~\cite{cao2026failures,qian2024streaming}. The adversary is an uploader who wants visually obvious harmful footage to pass the moderator while staying plainly recognizable to any human who watches it. It controls the uploaded file, so it may insert a short harmful clip into a benign carrier and apply a bounded perturbation $\delta$ with $\|\delta\|_\infty \le 8/255$ to that clip, while it may not touch the moderator's weights, its training data, the prompt issued to it, or any non-harmful frame of the benign carrier itself.

The adversary knows only that the moderator performs PGS-style selection, and not which method, which VLM scores frames, what $N$ is, or any other hyperparameter, and it cannot query the target while optimizing. Because PGS pipelines are assembled from publicly released components such as BLIP, CLIP, and small open VLMs, the adversary may optimize $\delta$ against a \emph{local} lightweight VLM and a public shadow \VideoLLM that need not match the target, which is the standard transfer setting~\cite{goodfellow2015explaining,madry2018pgd} and is what we mean by black-box. The attack succeeds when its output omits or misdescribes the inserted event, whether it was asked to detect harm or simply to describe the video, as we vary below.

\begin{figure*}[t]
    \centering
    \includegraphics[width=0.95\linewidth]{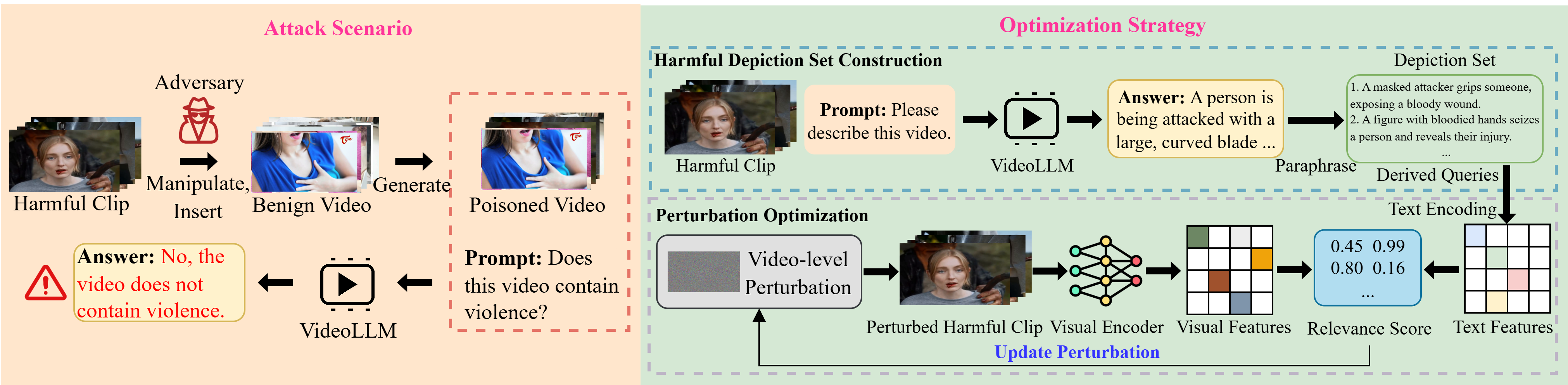}
    \caption{Overview of \ours. Left, the adversary inserts a harmful clip into a benign carrier so the \VideoLLM omits it. Right, \ours suppresses the prompt-conditioned relevance of that clip by optimizing against a depiction set drawn from it.}
    \label{fig:overview}
\end{figure*}

\subsection{Relevance Suppression}
\label{subsec:attack_formulation}

\Cref{fig:overview} summarizes the attack. Learning a perturbation per frame would be costly and unnecessary, so we optimize one video-level $\delta$ applied uniformly across the harmful clip $\mathcal{H}=\{f_1,\dots,f_h\}$, giving $\tilde f_j = f_j + \delta$. A naive objective would suppress $r(\tilde f_j, q)$ for the deployed moderator prompt $q$, and it fails twice, since the attacker never sees $q$, whose template is server-side, and since optimizing against one phrase binds the perturbation to text that may not be deployed.

We therefore optimize against a \emph{depiction set} $\mathcal{D}=\{d_1,\dots,d_l\}$ of natural-language descriptions of the harmful event. A shadow \VideoLLM captions the original clip to give a base description $\hat d$, GPT-4o-mini returns roughly 50 paraphrases of $\hat d$ at varied lengths and lexical styles, and we rank them by the cosine similarity of their BLIP text embeddings to $\hat d$, drop anything below 0.7 as drifted, and take the $l=5$ most similar survivors. The appendix reports what that threshold discards and how three human raters judge what it keeps. Each $d_k$ then becomes a relevance query $q'_k$ under the PGS template introduced above, collected together into $\mathcal{Q'}$. Scoring every perturbed frame against every one of these queries with a lightweight VLM such as BLIP~\cite{li2023blip} then yields the relevance suppression loss (RSL)
\begin{equation}
\label{eq:rsl_loss}
\mathcal{L}_{\mathrm{RSL}}(\delta;\mathcal{H},\mathcal{Q'})
= \frac{1}{hl} \sum_{j=1}^h \sum_{k=1}^l r(\tilde{f}_j,q'_k),
\end{equation}
which we minimize with respect to $\delta$ by projected gradient descent~\cite{madry2018pgd},
\begin{equation}
\label{eq:pgd}
\delta^{(t+1)} = \Pi_{\|\cdot\|_\infty \leq \epsilon} \Big( \delta^{(t)} - \eta_t \,\nabla_\delta \mathcal{L}_{\mathrm{RSL}}(\delta^{(t)}; \mathcal{H}, \mathcal{Q'}) \Big).
\end{equation}
Here $\Pi$ projects onto the $\ell_\infty$ ball of radius $\epsilon$ and the rate $\eta_t$ decays geometrically, so early steps move freely inside the ball and later ones settle, while steps follow the raw gradient rather than its sign and frames are clipped to the valid pixel range before scoring. Averaging over all of $\mathcal{Q'}$ rather than one phrase pushes $\delta$ to suppress the harmful \emph{semantics} instead of one phrase, which is what carries it to moderator prompts the attacker never saw. It does not carry across scoring backbones, and our analysis below shows why, while \Cref{alg:ours} states the procedure in full.

\section{Experiments}
\label{sec:experiment}

\subsection{Setup}\label{subsec:experiment_setup}

\noindent\textbf{Data.}
We draw 200 benign carriers at random from LLaVA-Video-178K \cite{zhang2024LLaVA-Video} and collect 60 harmful clips in six categories, namely \emph{violence}, \emph{crime}, \emph{porn}, \emph{terrorism}, \emph{scam}, and \emph{suicide}, with 10 clips each, sourced from public datasets such as RLVS~\cite{soliman2019violence} and XD-Violence~\cite{wu2020xd-violence} and from public online platforms.

\noindent\textbf{Models and Samplers.}
We evaluate six \VideoLLMs under each of DKS, AKS, and FRAG. Three are LLM-centric, namely LLaVA-Video-7B-Qwen2 (\LQwen)~\cite{zhang2024LLaVA-Video}, VideoLLaMA2 (\VideoLLaMA)~\cite{cheng2024videollama2}, and ShareGPT4Video (\SGV)~\cite{chen2024sharegpt4video}, and three are native multimodal, namely Qwen3-VL-8B-Instruct (\QwenVL)~\cite{yang2025qwen3}, InternVL3.5-8B (\InternVL)~\cite{wang2025internvl3}, and Molmo2-8B (\Molmo)~\cite{clark2026molmo2}. Proprietary models~\cite{hurst2024gpt4o,team2023gemini,team2024reka} are excluded because their reported pipelines still sample uniformly, which puts them outside the surface we target.

\noindent\textbf{Baselines and Metric.}
FRA~\cite{cao2026failures} is the only prior poisoning attack aimed at harmful-content omission and serves as the primary baseline under its own settings, joined by random Gaussian noise (RG) and FGSM~\cite{goodfellow2015explaining} to separate optimized relevance suppression from mere noise injection. We report Attack Success Rate (ASR), the fraction of poisoned videos on which the \VideoLLM fails to recognize the inserted harmful content. Each cell rests on 200 poisoned videos, so its standard error near 90\% is close to 2 points, while harmful clips are the scarcer axis at 10 per category. Every benign carrier is judged non-harmful by all six, and pre-attack recognition varies by model, so every comparison is read against the no-perturbation arm of that same model.

\noindent\textbf{Implementation.}
Following \cite{cao2026failures} we replace a contiguous 4-second segment of each carrier with a harmful clip, a duration shown to expose omission in minute-long video. \LQwen serves as the shadow \VideoLLM, GPT-4o-mini paraphrases, and BLIP scores relevance. We set $N=32$ for every sampler following mainstream defaults~\cite{li2024llava-onevision,zhang2024llavanextvideo} and modify only the sampling component. Perturbations obey $\|\delta\|_\infty \le \epsilon = 8/255$ from a uniform initialization~\cite{madry2018pgd,wong2020fast}, with distortion in \Cref{tab:imperceptibility} and \Cref{tab:quality_calibration}, and examples in \Cref{fig:examples}. We run 1{,}000 steps at initial rate 10 with decay 0.999, fix $|\mathcal{D}|=5$, and approximate \Cref{eq:rsl_loss} on 8 random harmful frames per iteration. We caption each harmful clip with \LQwen to standardize the sampler's scoring query, which tells PGS exactly what to look for and is the setting most favorable to it, and the pools for our minute-scale carrier videos each hold roughly 60 to 130 candidate frames, from which 32 are kept.

\noindent\textbf{Scope.}
\ours targets prompt-guided down-selection rather than \VideoLLMs with inherently dense context. Because the perturbation is applied to the inserted clip alone it is not tied to a particular budget and extends to 16, 64, or 128 frames (see the appendix). Every evaluated model therefore has its sampler replaced by the PGS method under test, including \QwenVL, whose native uniform sampling over up to 2{,}048 frames leaves no down-selection to attack.

\begin{table*}[!t]
\centering

\resizebox{0.95\textwidth}{!}{%
\def\arraystretch{.9}
\begin{tabular}{l|l|cccccc|cccccc|cccccc}
\toprule
\multirow{2}{*}{Category} & \multirow{2}{*}{Attack}
& \multicolumn{6}{c|}{DKS} & \multicolumn{6}{c|}{AKS} & \multicolumn{6}{c}{FRAG} \\
\cmidrule(lr){3-8} \cmidrule(lr){9-14} \cmidrule(lr){15-20}
 &  & \LQwen & \VideoLLaMA & \SGV & \QwenVL & \InternVL & \Molmo
 & \LQwen & \VideoLLaMA & \SGV & \QwenVL & \InternVL & \Molmo
 & \LQwen & \VideoLLaMA & \SGV & \QwenVL & \InternVL & \Molmo \\
\midrule
\multirow{4}{*}{Violence}
 & FRA        & 55 & 29 & 47 & 38 & 37 & 30 & 38 & 20 & 35 & 32 & 25 & 17 & 9 & 10 & 17 & 22 & 18 & 9 \\
 & RG & 43 & 26 & 41 & 31 & 36 & 35 & 40 & 24 & 30 & 27 & 22 & 13 & 8 & 8 & 10 & 19 & 20 & 5 \\
 & FGSM & 50 & 45 & 40 & 36 & 42 & 39 & 47 & 39 & 35 & 29 & 31 & 14 & 15 & 17 & 16 & 23 & 26 & 17 \\
\rowcolor{gray!5} & \ours      & 93 & 81 & 97 & 96 & 73 & 85 & 97 & 84 & 77 & 98 & 83 & 88 & 93 & 83 & 96 & 92 & 82 & 85 \\
\midrule
\multirow{4}{*}{Crime}
 & FRA        & 68 & 24 & 63 & 34 & 30 & 42 & 60 & 18 & 54 & 30 & 19 & 23 & 20 & 31 & 27 & 13 & 10 & 4 \\
 & RG & 55 & 33 & 56 & 29 & 36 & 37 & 52 & 14 & 51 & 32 & 22 & 16 & 17 & 10 & 26 & 9 & 13 & 5 \\
 & FGSM & 46 & 36 & 60 & 28 & 35 & 40 & 69 & 21 & 49 & 29 & 18 & 19 & 26 & 23 & 30 & 19 & 13 & 11 \\
\rowcolor{gray!5} & \ours      & 92 & 85 & 94 & 92 & 87 & 92 & 77 & 84 & 88 & 82 & 77 & 81 & 92 & 89 & 95 & 94 & 96 & 91 \\
\midrule
\multirow{4}{*}{Porn}
 & FRA        & 72 & 45 & 50 & 37 & 29 & 20 & 37 & 31 & 47 & 44 & 14 & 8 & 14 & 19 & 36 & 14 & 9 & 6 \\
& RG & 64 & 40 & 47 & 42 & 26 & 28 & 44 & 29 & 41 & 39 & 10 & 17 & 9 & 14 & 29 & 15 & 8 & 10 \\
 & FGSM & 60 & 33 & 47 & 40 & 26 & 25 & 48 & 34 & 46 & 42 & 11 & 14 & 21 & 7 & 29 & 17 & 13 & 19 \\
\rowcolor{gray!5} & \ours      & 94 & 85 & 98 & 96 & 91 & 72 & 91 & 88 & 95 & 96 & 80 & 83 & 74 & 81 & 95 & 96 & 96 & 95 \\
\midrule
\multirow{4}{*}{Terrorism}
 & FRA        & 47 & 48 & 44 & 39 & 38 & 47 & 31 & 32 & 34 & 36 & 25 & 24 & 11 & 21 & 18 & 28 & 10 & 13 \\
 & RG & 44 & 51 & 43 & 39 & 33 & 43 & 37 & 30 & 39 & 30 & 16 & 19 & 10 & 22 & 14 & 22 & 9 & 16 \\
 & FGSM & 58 & 43 & 46 & 42 & 40 & 52 & 38 & 29 & 37 & 31 & 19 & 22 & 14 & 15 & 19 & 25 & 6 & 12 \\
\rowcolor{gray!5} & \ours      & 97 & 80 & 99 & 98 & 93 & 83 & 95 & 78 & 98 & 89 & 86 & 78 & 96 & 76 & 98 & 92 & 86 & 81 \\
\midrule
\multirow{4}{*}{Scam}
 & FRA        & 39 & 50 & 41 & 43 & 28 & 51 & 30 & 32 & 25 & 33 & 27 & 32 & 17 & 16 & 20 & 22 & 15 & 15 \\
 & RG & 46 & 50 & 40 & 38 & 37 & 48 & 37 & 34 & 23 & 35 & 24 & 29 & 20 & 7 & 15 & 25 & 20 & 8 \\
 & FGSM & 43 & 53 & 43 & 40 & 32 & 50 & 35 & 31 & 30 & 30 & 26 & 24 & 14 & 12 & 22 & 23 & 17 & 14 \\
\rowcolor{gray!5} & \ours      & 87 & 84 & 96 & 96 & 89 & 84 & 98 & 98 & 99 & 95 & 90 & 89 & 93 & 90 & 97 & 89 & 93 & 86 \\
\midrule
\multirow{4}{*}{Suicide}
 & FRA        & 46 & 34 & 40 & 38 & 31 & 62 & 34 & 26 & 37 & 29 & 22 & 22 & 13 & 7 & 34 & 12 & 21 & 19 \\
 & RG & 47 & 32 & 37 & 40 & 29 & 55 & 29 & 28 & 32 & 22 & 16 & 31 & 11 & 10 & 30 & 16 & 14 & 26 \\
 & FGSM & 49 & 37 & 39 & 41 & 35 & 58 & 30 & 31 & 32 & 26 & 17 & 25 & 18 & 16 & 28 & 17 & 18 & 21 \\
\rowcolor{gray!5} & \ours      & 96 & 86 & 97 & 98 & 90 & 97 & 92 & 90 & 98 & 87 & 87 & 94 & 98 & 88 & 99 & 90 & 89 & 92 \\
\midrule
\multirow{4}{*}{Average}
 & FRA        & 55 & 38 & 48 & 38 & 32 & 42 & 38 & 27 & 39 & 34 & 22 & 21 & 14 & 17 & 25 & 19 & 14 & 11 \\
 & RG & 50 & 39 & 44 & 37 & 33 & 41 & 40 & 27 & 36 & 31 & 18 & 21 & 13 & 12 & 21 & 18 & 14 & 12 \\
 & FGSM & 51 & 41 & 46 & 38 & 35 & 44 & 45 & 31 & 38 & 31 & 20 & 20 & 18 & 15 & 24 & 21 & 16 & 16 \\
\rowcolor{gray!5} & \ours      & 93 & 84 & 97 & 96 & 87 & 86 & 92 & 87 & 93 & 91 & 84 & 86 & 91 & 85 & 97 & 92 & 90 & 88 \\
\bottomrule
\end{tabular}%
}
\caption{ASR (\%) of FRA \cite{cao2026failures}, random Gaussian (RG), FGSM~\cite{goodfellow2015explaining}, and \ours against three PGS methods on six \VideoLLMs. Higher is better for the attacker. Every cell is measured on 200 poisoned videos built from the 10 harmful clips of its category, and the shaded row gives \ours in each of the blocks.}
\label{tab:main_results}
\end{table*}

\subsection{Attack Effectiveness}

\Cref{tab:main_results} reports the headline comparison. \ours outperforms FRA on every one of the 18 sampler and model pairs, with average ASR between 84\% and 97\%, while Gaussian noise and FGSM are ineffective and sometimes fall below FRA, so success requires a perturbation optimized for the task and not merely energy spent inside the pixel budget. Per-cell values run from 72 to 99, so not one combination of sampler, model, and harmful category escapes the attack.

Under FRA the three samplers order themselves exactly as their designs predict. DKS suppresses redundancy and so keeps only a few harmful frames, AKS spends part of its budget on temporal coverage, and FRAG ranks by relevance alone, so harmful frames that score highly are selected regardless of redundancy, which makes FRAG the strongest of the three and holds FRA to 16.7\%. Relevance is therefore the one quantity that separates them, and it is the quantity \ours removes, driving the harmful-frame relevance from its pre-attack range of 0.81 to 0.99 down to below 0.001.

Transfer holds along both axes that matter to an attacker who cannot query the target. Five of six models are not the shadow \VideoLLM, and \ours reaches 84 to 97 average ASR on each, so it never depended on seeing its target. BLIP scores relevance during optimization, and the perturbation still defeats samplers scored by a different VLM, as \Cref{tab:ablation_vlm} shows for each backbone.

\subsection{Failure at the Sampling Stage}
\label{subsec:mechanism}

A relevance-suppressing perturbation may evict harmful frames from the selection, or it may damage them so the \VideoLLM no longer reads them as harmful, which would make \ours an ordinary attack on the encoder. Comparing samplers that read the score against samplers that ignore it separates the two. Throughout this subsection one carrier and one insertion point are held fixed while only the perturbation varies, and we report harmful frames retained out of the 4 that a 4-second clip contributes at 1 FPS, with the protocol set out in full in the appendix.

\noindent\textbf{Samplers That Read the Score.}
The sampling account makes a falsifiable prediction, since a perturbation that damaged the frames would cost them their place under a sampler that never scores relevance. No such cost appears over 48 paired videos, where UFS retains 0.92 harmful frames with and without the perturbation and SSS is unchanged within noise at 1.52 against 1.44, while AKS, which ranks by the very score under attack, falls from 1.69 to 0.71 on those same videos, so what changes tracks what a sampler reads.

\noindent\textbf{Scoring Backbone and Evictability.}
Suppressing a score helps only where the sampler decides by it. We optimize a second perturbation against CLIP, the backbone DKS reads, and run both on 36 paired videos. Each moves only its own backbone, since the BLIP-targeted one drives the BLIP score from 0.994 to 0.0002 and leaves CLIP at 0.285, while the CLIP-targeted one drives CLIP from 0.287 to $-0.286$ and leaves BLIP at 0.993. Against AKS the matched perturbation removes 64\% of the harmful frames it would otherwise keep and the mismatched one removes none, while against DKS the matched one removes 13\%, because DKS spreads its remaining picks over the timeline once its score-driven pass ends. Each perturbation drove its own sampler's scorer to the floor, so what separates 64\% from 13\% is the selection rule rather than the VLM underneath it. A sampler is evictable to the extent that its output is decided by the signal under attack, which is the practical lesson here for a defender.

\noindent\textbf{Prompts the Attacker Never Saw.}
The depiction set buys independence from the moderator's phrasing, so we test it on prompts withheld from optimization. Over 72 paired videos it evicts 54\% of the harmful frames AKS would otherwise keep against 49\% for a perturbation fitted to a single description, and a plain question no part of the attack ever saw, such as whether the video contains any violence, raises eviction to 73\%, so the set generalizes past its own wording.

\noindent\textbf{Encoder Attacks at Equal Budget.}
Perturbing a vision-language embedding is well studied, so the depiction set has to earn its place against that alternative at the same budget on the same videos. A text-free feature-collapse attack, which pushes each frame embedding off its clean position, drives relevance to 0.003 and evicts a third of the harmful frames, whereas relevance suppression reaches 0.0003 and evicts over half. Attacking the encoder gets part of the way and aiming at what the sampler reads goes much further, while uniform sampling gives an identical 87.5\% for every one of them, since it reads no score at all.

\noindent\textbf{Harmful-Frame Count.}
The same picture holds at scale in \Cref{fig:selected_harmful_frames_all}, where under FRA, AKS and FRAG select between 3.5 and 3.8 of the 4 available harmful frames, which shows how aggressively prompt guidance favors harmful content once a query names it, and under \ours the count falls to roughly 1.0 to 1.5. That residual reflects saturation of the top-$N$ rule rather than failed suppression, since a pool of 60 to 130 candidates must still fill 32 slots, and those few survivors cannot dominate the response~\cite{cao2026failures}.

\subsection{Prompt Variation, Ablation, and Transfer}
\label{subsec:prompts}

\noindent\textbf{Prompt Regimes.}
To avoid tying our conclusions to one prompt, we also evaluate description, summarization, and a multi-hop dialogue that asks the model to detect, describe, then re-check after guidance (\Cref{tab:more_prompts_all}). Multi-hop needs multi-turn dialogue, which \LQwen, \VideoLLaMA, and \SGV lack. Summarization prompts often yield higher ASR, and explicit guidance in the multi-hop setting leaves ASR at or above the single-turn level, so telling the model to look again cannot recover evidence never sampled. It sometimes concedes harmfulness yet misidentifies the event, reporting a shooting as a theft, which is the very same failure seen at the output.

\noindent\textbf{Relevance-Scoring Backbone.}
\label{subsec:ablation_study}
\Cref{tab:ablation_vlm} varies the surrogate used during optimization over BLIP, CLIP, and their average on \LQwen. Since the three samplers score with three different backbones, this also tests whether the attack must match the deployed scorer. BLIP transfers uniformly, averaging 93.2 on DKS, 91.7 on AKS, and 91.0 on FRAG, and AKS is the one sampler that actually uses BLIP yet sits between the two that do not, so BLIP loses nothing by being unmatched. CLIP averages 91.8 on the CLIP-based DKS while falling to 73.0 on AKS and 76.3 on FRAG, and the combined variant inherits that weakness. CLIP is thus a narrow surrogate whose strength follows the backbone and BLIP a broad one. An adversary who cannot query the target cannot run this comparison, so the choice must rest on how widely a scorer is deployed, which is what makes BLIP the right default here.

\noindent\textbf{Harmful Clip Length.}
Longer clips are more salient and raise the chance that harmful frames survive suppression, and \Cref{fig:ablation_length_all} confirms that ASR falls with clip length under all three samplers, steeply for FRAG and AKS and gently for DKS. Even at 36 seconds \ours still averages 68\% under DKS, so length blunts the threat without removing it.

\begin{table}[t]
\centering

\resizebox{\columnwidth}{!}{%
\def\arraystretch{.9}
\begin{tabular}{lcccccc}
\toprule
\diagbox{Strategy}{Model} & \LQwen & \VideoLLaMA & \SGV & \QwenVL & \InternVL & \Molmo \\
\midrule
UFS          & 97 & 97 & 99 & 97 & 93 & 96 \\
\midrule
SSS: SKE      & 94 & 89 & 98 & 96 & 92 & 91 \\
\midrule
PGS: DKS      & 93 & 84 & 97 & 96 & 87 & 86 \\
PGS: AKS      & 92 & 87 & 93 & 91 & 84 & 86 \\
PGS: FRAG     & 91 & 85 & 97 & 92 & 90 & 88 \\
\bottomrule
\end{tabular}
}
\caption{ASR (\%) of \ours under five sampling strategies, averaged over the six harmful categories.}
\label{tab:other_sampling}
\end{table}

\noindent\textbf{Transfer to Prompt-Agnostic Samplers.}
\Cref{tab:other_sampling} runs \ours against UFS and SSS as well, and the high ASR it reaches there is a natural transfer rather than a second mechanism. Neither sampler reads the query, so neither can aim at a four-second insert, and \Cref{fig:pre-example} shows both forwarding a single harmful frame under plain frame replacement alone. A perturbation that lowers relevance cannot change what a sampler that never computes relevance decides to keep, so on UFS and SSS \ours simply inherits the omission that FRA already produced. That column matters on its own for deployment, because UFS remains the default in most released \VideoLLMs, and 93 to 99 there says a poisoned upload passes unreported through the pipeline that platforms most often run today, with the event left plainly visible to any human who watches the file. PGS is the sampler on which the perturbation has to earn its result, and it stays the hardest of the five at 88.8 to 90.5 average ASR against 96.5 for UFS. What \ours changes is the size of that gap, since frame replacement costs an attacker 48 to 73 points on PGS relative to UFS while \ours cuts the price to 6 to 8 points.

\begin{table}[t]
\centering

\resizebox{\columnwidth}{!}{%
\def\arraystretch{.9}
\begin{tabular}{lcccccc}
\toprule
\diagbox{Defense}{Category} & Violence & Crime & Porn & Terrorism & Scam & Suicide \\
\midrule
None                                & 93 & 92 & 94 & 97 & 87 & 96 \\
\midrule
AD: PAD~\cite{jing2024pad}          & 90 & 87 & 90 & 97 & 88 & 94 \\
AD: AdvIT~\cite{xiao2019advit}        & 95 & 91 & 92 & 94 & 85 & 90 \\
AD: Comdefend~\cite{jia2019comdefend}    & 86 & 90 & 88 & 93 & 83 & 84 \\
\midrule
DPD: CD~\cite{steinhardt2017certified}          & 86 & 81 & 89 & 80 & 91 & 86 \\
DPD: De-Pois~\cite{chen2021depois}     & 84 & 80 & 84 & 86 & 87 & 86 \\
\midrule
ITD: VLM-Det & 79 & 88 & 86 & 90 & 81 & 87 \\
ITD: ERS          & 75 & 78 & 73 & 83 & 74 & 82 \\
\bottomrule
\end{tabular}
}
\caption{ASR (\%) of \ours under three categories of defenses, measured on \LQwen under DKS and averaged over all six of the harmful categories in the row for no defense.}
\label{tab:defenses}
\end{table}

\subsection{Defenses and Mitigations}

\Cref{tab:defenses} evaluates seven defenses in three families, adversarial (AD), data-poisoning (DPD), and inference-time (ITD), each applied before sampling so the comparison isolates what the defense recovers. From an undefended average of 93.2, the purification defenses PAD~\cite{jing2024pad}, AdvIT~\cite{xiao2019advit}, and ComDefend~\cite{jia2019comdefend} cost between 2.0 and 5.9 points. Their failure is structural. Because $\delta$ is identical on every frame it moves with the scene and leaves frame-to-frame differences intact, which is what AdvIT inspects through optical flow, while the artifact removal of PAD and ComDefend cannot localize a perturbation that is spread evenly across the whole frame.

The data-poisoning detectors CD~\cite{steinhardt2017certified} and De-Pois~\cite{chen2021depois} target manipulated training data and cost 7.7 and 8.7 points. \ours touches neither training data nor parameters, so what either of them removes here is a side effect of distribution filtering rather than of real detection.

The two inference-time mitigations are more informative, since VLM-Det adds an auxiliary harm detector over the sampled frames and costs 8.0 points, while ERS aggregates relevance scores across several lightweight VLMs and costs 15.7, which makes it the most effective of the seven and the only one to exceed ten points. ERS works for the reason the backbone experiment predicts, since a perturbation tuned to one scorer leaves another untouched. Individual cells move a point or two either way, including one where a defense raises ASR, which a standard error near 2 points predicts.

\noindent\textbf{Test-Time Counterattacks.}
A recent defense line purifies an input at inference by counterattacking it, where TTC~\cite{xing2025ttc} takes a few ascent steps that push a CLIP image embedding away from where the input places it, since an adversarial example sits in a shallower basin than a clean one. The step drops in directly before a relevance scorer, so the family applies to our pipeline and we did not evaluate it. Relevance suppression drives the matching head it targets from 0.994 to 0.0002 while leaving CLIP at 0.285 against 0.287 clean, so TTC in its published CLIP form has little to reverse and a defender would have to port the counterattack onto whichever head the sampler scores with, at several extra passes over a pool of 60 to 130 candidates. That cost and robust fine-tuning of the scorer~\cite{schlarmann2024fare} both belong in a sampling-stage comparison we leave to follow-up rather than settle here.

\noindent\textbf{Practitioner Guidance.}
Four measures follow from the analysis and need no change to the models. Randomize the selection budget, so an attacker cannot know how far a score must be pushed. Run the harm filter on a backbone the relevance scorer does not share, which is what gives ERS its effect. Audit a random sample of the discarded frames, so a suppressed region leaves a trace. Log the relevance scores, so an abrupt collapse is visible afterwards. None of them suffices alone in \Cref{tab:defenses}, and we offer them as cost-bearing mitigations rather than solutions to the underlying flaw.

\noindent\textbf{Adaptive Adversaries.}
A defense is meaningful only against an adaptive adversary, so the non-adaptive numbers in \Cref{tab:defenses} upper-bound the residual robustness of ERS and VLM-Det rather than establishing it. An adaptive attacker can add ERS's auxiliary VLMs or VLM-Det's classifier to $\mathcal{Q'}$ and close the transferability gap this evaluation exploits. What remains is therefore an artifact of transfer and not a structural barrier, which implies that no defense confined to the relevance-scoring layer can hold. Stronger directions intervene outside that layer, by randomized over-sampling with VLM-independent harm filtering, temporal-consistency checks on adjacent relevance scores, or certified scoring through randomized smoothing. Each costs computation and eats into the efficiency motivating PGS.

\begin{table}[t]
\centering

\setlength{\tabcolsep}{6pt}
\resizebox{\columnwidth}{!}{%
\def\arraystretch{.95}
\begin{tabular}{lccc}
\toprule
\textbf{Ingestion} & \textbf{Relevance} & \textbf{Harmful frames} & \textbf{ASR (\%)} \\
\midrule
None            & 0.0002 & 0.72 & 94 \\
H.264 CRF 28    & 0.018  & 1.22 & 83 \\
H.265 CRF 28    & 0.049  & 1.28 & 82 \\
H.264 CRF 32    & 0.061  & 1.52 & 77 \\
Downscale 360p  & 0.482  & 1.85 & 46 \\
\bottomrule
\end{tabular}
}
\caption{\ours under platform-style transcoding on AKS, measured at selection and at the output.}
\label{tab:ingestion}
\end{table}

\subsection{Robustness to Platform Ingestion}
\label{subsec:ingestion}

A moderator never sees the file the adversary uploaded, because platforms re-encode on ingest and usually emit several resolutions. \Cref{tab:ingestion} therefore transcodes each poisoned video and re-runs both selection and the moderator on what comes out. Every transcode gives back part of what the perturbation took, and the more heavily it resamples the more it gives back. H.264 at CRF 28 lifts harmful-frame relevance from 0.0002 to 0.018 and the count from 0.72 to 1.22, which costs 11 points of ASR, H.265 at the same quality lands within a point of it at 82, CRF 32 lifts relevance to 0.061 and returns 1.52 frames at 77, and downscaling to 360p goes furthest at 0.482 relevance and 1.85 frames, which leaves ASR at 46. Selection and output move together throughout, and resampling blunts \ours by 48 points without ever closing the surface that prompt-guided selection opens.

The pattern follows from what selection depends on, since a frame is kept or dropped by its rank, so the perturbation need only lose its margin over competing frames to stop working and never has to be erased. Resampling costs the attack more than quantization does, because the perturbation lives on the pixel grid and averaging neighboring pixels damages it, yet even a 360p ladder leaves most of the ranking poisoned. An adversary facing a real ladder could recover the rest by optimizing through it, taking the expectation of \Cref{eq:rsl_loss} over the codecs and scales it applies.

\section{Conclusion}
\label{sec:conclusion}

We have examined the safety of prompt-guided sampling and found its defining strength to be its exposure. \ours is the first black-box attack on that surface, poisoning the sampler's ranking with one perturbation anchored on a depiction set of paraphrased harmful descriptions. It needs no access to the moderator's weights or to the prompt it will be asked, and it carries to samplers and models the attacker never saw.

The failure sits in the sampler rather than in the model that answers. Samplers blind to the score keep the frames they always kept, and only the one that ranks by the score under attack loses them, so hardening the encoder or the language model gains nothing, because the evidence is gone before either runs. Selection therefore acts as a security boundary wherever a moderator ranks by prompt relevance. Every mitigation that helps at all works by not trusting that score, and each spends the computation prompt-guided sampling exists to save. Deciding how much of that saving to give back is where the next defenses begin.

\clearpage
\bibliography{refs}

\clearpage
\section{Appendix}
\label{sec:appendix}

\subsection{Attack Algorithm}
\label{subsec:algorithm}
\Cref{alg:ours} states the optimization procedure of \ours, which initializes the perturbation, derives the query set from the depiction set, and then iteratively suppresses the relevance score of the clip subject to an $\ell_\infty$ projection on it.

\begin{algorithm}[h]
\small
\caption{Optimization procedure of \ours}
\label{alg:ours}
\textbf{Input}: Harmful clip $\mathcal{H}=\{f_1,\dots,f_h\}$, depiction set $\mathcal{D}=\{d_1,\dots,d_l\}$, initial perturbation $\delta_0$, learning rate $\eta$, perturbation constraint $\epsilon$, epochs $Z$.\\
\textbf{Output}: Video-level perturbation $\delta$.
\begin{algorithmic}[1]
\STATE $\delta \leftarrow \delta_0$
\STATE Derive query set $\mathcal{Q'}=\{q'_1, q'_2, \dots, q'_l\}$ from $\mathcal{D}$
\FOR{$epoch = 1$ {\bfseries to} $Z$}
  \STATE Apply perturbation: $\tilde{f}_j \leftarrow \mathrm{clip}_{[0,1]}(f_j+\delta)$ for $j=1,\dots,h$
  \STATE Compute relevance suppression loss:
  \STATE \quad $\mathcal{L}_{\mathrm{RSL}}(\delta;\mathcal{H},\mathcal{Q'}) = \frac{1}{hl} \sum_{j=1}^h \sum_{k=1}^l r(\tilde{f}_j,q'_k)$
  \STATE Update perturbation: $\delta \leftarrow \delta - \eta_{epoch} \nabla_\delta \mathcal{L}_{\mathrm{RSL}}$, with $\eta_{epoch}=\eta\gamma^{epoch}$
  \STATE Project: $\delta \leftarrow \Pi_{\|\cdot\|_\infty \leq \epsilon}(\delta)$
\ENDFOR
\STATE \textbf{return} $\delta$
\end{algorithmic}
\end{algorithm}

\subsection{Sampling Methods in Detail}
\label{sec:sampling_detail}

\noindent\textbf{The Samplers We Attack.}
The three PGS implementations we attack differ only in what they add above relevance scoring. DKS~\cite{cheng2025vilamp} samples at 1 FPS, scores frames against the query with CLIP~\cite{radford2021clip}, and penalizes feature similarity between a frame and its neighbors when forming the final set. AKS~\cite{tang2025aks} also pools at 1 FPS but scores with a lightweight VLM such as BLIP~\cite{li2023blip}, where the score reflects whether a frame carries enough information to answer the query, and it adds a coverage term that spreads the selection over the timeline. FRAG~\cite{huang2025frag} uniformly samples 256 frames, scores each with a full \VideoLLM, and keeps the top 32 by relevance alone, with no term for redundancy or for coverage.

\noindent\textbf{Related Prompt-Aware Selectors.}
Two further selectors rank on the prompt without forming a family of their own. Frame-Voyager~\cite{yu2025framevoyager} learns a query-aware ranker over frame subsets, but it is supervised by a single \VideoLLM and restricted to short videos, and FocusChat~\cite{cheng2024focuschat} applies a spatial-temporal filter over EVA-CLIP features in order to discard the frames that are irrelevant to the prompt. Both decide admission by a learned relevance signal, so both inherit the exposure this paper measures on DKS, AKS, and FRAG, and neither one is evaluated here.

\noindent\textbf{Other Sampling Strategies.}
Beyond the three families this paper treats, several sampling approaches have emerged that have not yet formed distinct or widely adopted categories. VideoAgent~\cite{fan2024videoagent} samples through an LLM-guided multi-round process, first taking a small uniform set for context, then letting the LLM identify missing information and name the segment to explore next, and finally retrieving frames within that segment by CLIP relevance. Flow4Agent~\cite{liu2025flow4agent} uses coarse optical flow to segment the video into dynamic events and applies semantic filtering at the centers of those events to pick representative keyframes. Since our attack is designed for PGS, extending the analysis to these isolated and still evolving approaches lies beyond the present scope and is left to future work here.

\subsection{More Experimental Details}\label{sec:more_setup}

\noindent\textbf{Semantic Control.}
When constructing the depiction set, we first query the paraphrasing language model with instructions such as ``Please provide some paraphrasing versions of the following sentence which depicts a violent scene from a video. You can provide sentences with different lengths. This video shows ...'' to obtain a large pool of paraphrased sentences. We then apply a cosine-similarity filter on BLIP text embeddings to control semantic deviation, which removes sentences whose meaning drifts too far from the original description and would otherwise pull the optimization toward an event that the clip does not show at all.

\noindent\textbf{Gradient Source.}
Gradients are read on the matching logit before its final normalization, since a clip that matches its depiction set strongly sits at the ceiling of the normalized score and carries almost no gradient there.

\noindent\textbf{Computation Cost.}
During perturbation optimization, each iteration takes an average of 5.8 seconds on a single NVIDIA RTX 6000 Ada GPU when relevance is scored one frame-text pair at a time, and about 0.5 seconds once the pairs are batched. The loss flattens within 100 to 180 iterations, so the 1{,}000-step budget is conservative. Running it in full costs about 8 minutes at the batched rate, and stopping at convergence brings that under 2 minutes. The optimization runs offline and adds no latency at deployment time.

\subsection{Depiction Set Quality}
\label{sec:depiction_quality}

Two models and one threshold decide the depiction set, so this section reports what the threshold discards and what three human raters say about the descriptions it keeps.

\noindent\textbf{What the Threshold Discards.}
The filter ranks the roughly 50 paraphrases of a clip by the cosine similarity of their BLIP text embedding to the base description, drops anything below $\tau=0.7$, and keeps the 5 most similar survivors. \Cref{tab:threshold} reports what that does over 260 candidates drawn from one clip per category. Their median similarity is 0.953, the filter discards no candidate at all up to 0.7, and the 5 descriptions the attack actually uses sit at 0.964 or above on every clip. Moving $\tau$ anywhere between 0 and 0.96 therefore leaves every depiction set unchanged, and with it the perturbation and every number computed from it. Only above 0.96 does the filter begin to bite, where clips start returning fewer than 5 descriptions and the depiction set collapses toward the single base description. The threshold is a guard against a failure mode this paraphraser does not produce rather than a value the attack was tuned on.

\begin{table}[t]
\centering

\setlength{\tabcolsep}{5pt}
\resizebox{\columnwidth}{!}{%
\def\arraystretch{.95}
\begin{tabular}{lccc}
\toprule
Threshold $\tau$ & Candidates kept (\%) & Kept per clip & Sets unchanged \\
\midrule
0.50 & 100.0 & 43.3 & 6/6 \\
0.70 & 100.0 & 43.3 & 6/6 \\
0.80 & 98.1  & 42.5 & 6/6 \\
0.90 & 91.4  & 39.7 & 6/6 \\
0.95 & 52.3  & 22.7 & 6/6 \\
0.96 & 41.5  & 18.0 & 6/6 \\
0.97 & 26.2  & 11.5 & 5/6 \\
0.98 & 13.7  & 5.8  & 2/6 \\
\bottomrule
\end{tabular}
}
\caption{Effect of the similarity threshold, over 260 paraphrase candidates from one clip per category.}
\label{tab:threshold}
\end{table}

\noindent\textbf{Human Validation.}
We check both stages by hand, where three of the authors rate independently and do not confer, and they are the only people who see the harmful material, for the reason given under ethical considerations below. The first stage asks whether the base description names the event the clip actually shows, which needs the clip and is rated on all 60 of them. The second stage asks whether a description depicts the same event as the base description, which needs only the text. For that stage each clip contributes the 5 depictions the attack uses, the 5 lowest-scoring candidates from its pool, and 5 rewrites we obtain by asking the paraphrasing model to change the event rather than preserve it, which supplies the low end of the similarity axis the paraphraser does not otherwise reach. The 900 items are pooled, shuffled, and shown without their similarity or their source, so a rater cannot tell a retained description from a drifted one.

\Cref{tab:human} gives the outcome, where the base description names the event correctly on 56 of the 60 clips and the four it misses name a related act instead, such as a robbery for an assault. Among the 5 depictions the attack uses, 97.2\% are judged to depict the same event, and the 5 lowest-scoring candidates reach 93.7\%, so the pool the paraphraser returns is already almost clean before any filtering. The deliberately drifted rewrites fall to 12.6\% at a mean similarity of 0.61. Agreement across the three raters is a Fleiss' $\kappa$ of 0.81, and read as a classifier of the human majority the threshold at 0.7 is right on 94.4\% of the 900 items.

\begin{table}[t]
\centering

\setlength{\tabcolsep}{5pt}
\resizebox{\columnwidth}{!}{%
\def\arraystretch{.95}
\begin{tabular}{lccc}
\toprule
Item group & Items & Mean similarity & Same event (\%) \\
\midrule
Base description against clip & 60  & --   & 93.3 \\
Retained depictions           & 300 & 0.98 & 97.2 \\
Lowest-scoring candidates     & 300 & 0.87 & 93.7 \\
Drifted rewrites              & 300 & 0.61 & 12.6 \\
\bottomrule
\end{tabular}
}
\caption{Human validation of the depiction sets, rated independently by three authors.}
\label{tab:human}
\end{table}

\noindent\textbf{Similarity and Semantic Correctness.}
The two stages together say what the cosine number does and does not mean. On its own it is no measure of semantic correctness, since a paraphrase scores 0.98 against a base description that was itself wrong about the clip, which is what the four ungrounded captions show. What it measures is agreement with the base description, and at that it separates cleanly, because same-event items concentrate above 0.84 while the drifted rewrites fall below 0.7 and almost nothing lands in between. The threshold sits in that empty band, which is why its exact value does not matter. Correctness has to be established at the captioning stage instead, where similarity has nothing to say and where the human check above supplies the evidence that no similarity number can give.

\subsection{Selection-Stage Measurements in Full}
\label{sec:mechanism_detail}

This section sets out the selection-stage measurements of the main text in full, and every one of them follows the same protocol. A carrier and an insertion position are fixed, only the perturbation varies between arms, and we report a harmful-frame count out of 4, since a 4-second clip contributes 4 frames at 1 FPS. The experiments draw different subsets of carriers, so the number of pairs is stated with each result and the no-perturbation baseline sits at 1.69 across them. These counts sit below the 3.5 to 3.8 that \Cref{fig:selected_harmful_frames_all} reports under FRA, because our carriers give AKS a pool of about 136 candidates for 32 slots while the main experiments use pools of 60 to 130, and a sampler admitting a quarter of its pool keeps fewer harmful frames than one admitting half. Every comparison is between arms measured on the same videos, which the paired design guarantees.

\noindent\textbf{Attacks at Equal Budget.}
We compare five perturbations at the same $\ell_\infty$ budget on the same videos and read the effect at the selection stage. Random Gaussian noise and one-step FGSM leave the relevance of the harmful frames at 0.99 and 0.93 and remove none of them from the selection, which reproduces the baseline behavior in \Cref{tab:main_results}. A text-free feature-collapse attack, which pushes each frame embedding away from its clean position on the same backbone, drives relevance to 0.003 and removes a third of the harmful frames. The relevance-suppression objective drives relevance to 0.0003 and removes over half. %
Under uniform sampling all six perturbations give the identical output rate of 87.5\% on \QwenVL, which is what a sampler that never scores a frame has to give. The other output cells hold 24 pairs each, so a difference there carries a standard error near 12 points, and we read the comparison between perturbations from the selection stage instead of the output.

\noindent\textbf{Prompts the Attacker Never Saw.}
We hold out the descriptions used for optimization and drive the sampler with four prompts the perturbation never saw, on 72 paired videos per arm. Against AKS the depiction set evicts 54\% of the harmful frames AKS would otherwise keep, and a perturbation optimized against one description evicts 49\%. The ordering holds on all four prompts and the depiction set varies less across them, with a standard deviation of 7.6 points against 8.6. Against DKS neither perturbation evicts anything at any prompt, whichever description drives the sampler.

A stronger version of the test removes the shared generator entirely. Both the depiction set and the scoring query above are written by the same shadow \VideoLLM from the same clip, so agreement between them could reflect a shared style rather than genuine transfer. We therefore drive the sampler with a plain question no part of the attack ever saw, for instance whether the video contains violence. Against AKS the perturbation evicts 73\% of the harmful frames over 36 pairs, above the 61\% it reaches when the sampler scores against a caption of the clip itself. The generic question separates frames less sharply, so harmful frames start with a smaller margin over the rest of the pool and less suppression is needed to push them out, which is why the number rises rather than falls once the query stops naming the event.

\subsection{More Experimental Results}\label{sec:more_results}

\begin{figure*}[!t]
    \centering
    \includegraphics[width=0.9\linewidth]{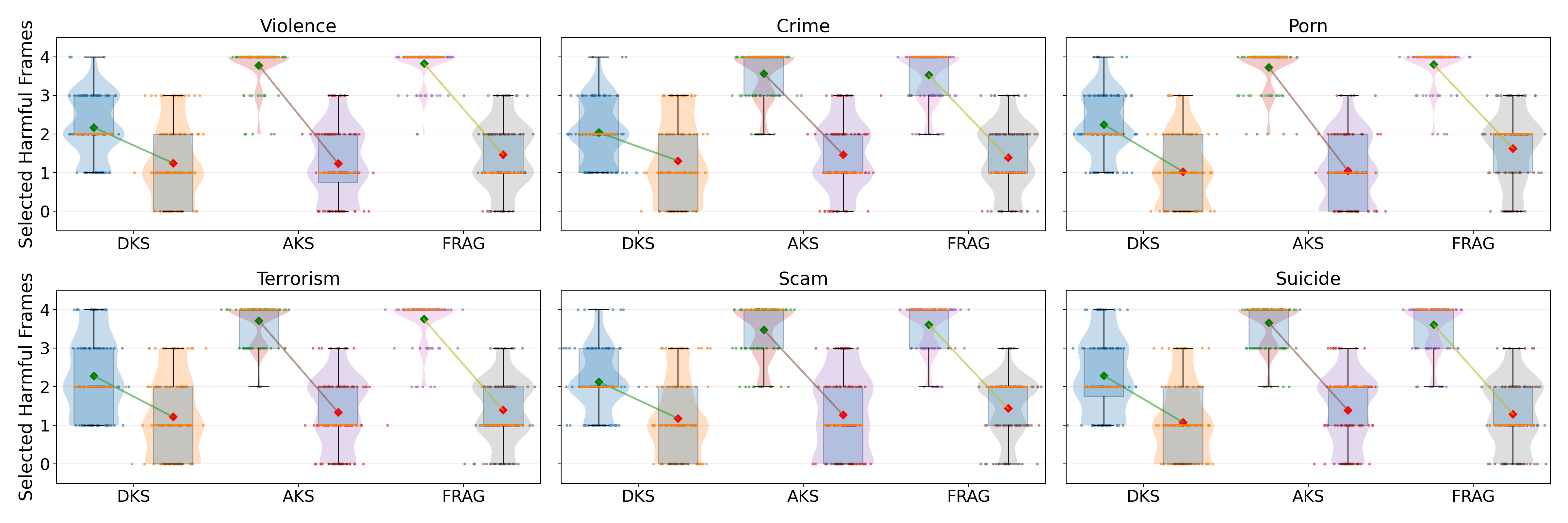}
    \caption{Selected harmful-frame counts before poisoning (\ie FRA, green diamonds) and after poisoning (\ie \ours, red diamonds) across all six harmful categories.}
    \label{fig:selected_harmful_frames_all}
\end{figure*}

\begin{figure*}[!ht]
    \centering
    \includegraphics[width=0.95\linewidth]{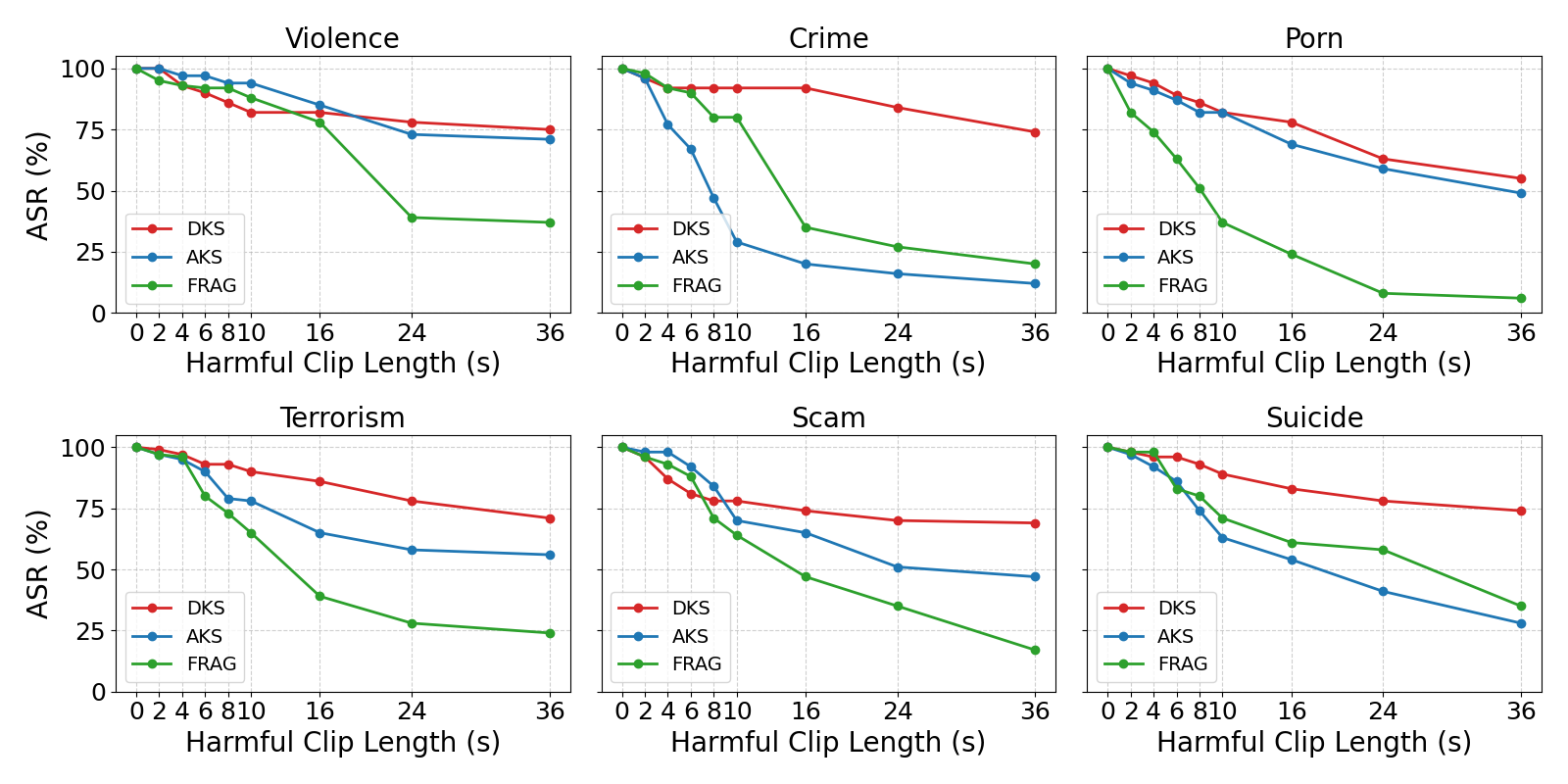}
    \caption{Attack performance across harmful clip lengths for all six harmful categories.}
    \label{fig:ablation_length_all}
\end{figure*}

For space, the main paper summarizes several experiments and defers their tables here. \Cref{fig:selected_harmful_frames_all} gives harmful-frame counts, \Cref{tab:more_prompts_all} the four prompt regimes, \Cref{tab:ablation_vlm} the relevance-scoring backbones, and \Cref{fig:ablation_length_all} the clip-length sweep, in each case measured across all six of the harmful categories rather than the subset shown in the main paper.

\begin{table*}[!t]
\centering

\def\arraystretch{.9}
\resizebox{\textwidth}{!}{%
\begin{tabular}{l|l|cccccc|cccccc|cccccc}
\toprule
\multirow{2}{*}{Category} & \multirow{2}{*}{Setting}
& \multicolumn{6}{c|}{DKS} & \multicolumn{6}{c|}{AKS} & \multicolumn{6}{c}{FRAG} \\
\cmidrule(lr){3-8} \cmidrule(lr){9-14} \cmidrule(lr){15-20}
 &  & \LQwen & \VideoLLaMA & \SGV & \QwenVL & \InternVL & \Molmo
 & \LQwen & \VideoLLaMA & \SGV & \QwenVL & \InternVL & \Molmo
 & \LQwen & \VideoLLaMA & \SGV & \QwenVL & \InternVL & \Molmo \\
\midrule
\multirow{4}{*}{Violence}
 & Prompt-1      & 93 & 81 & 97 & 96 & 73 & 85 & 97 & 84 & 77 & 98 & 83 & 88 & 93 & 83 & 96 & 92 & 82 & 85 \\
 & Prompt-2      & 95 & 90 & 98 & 97 & 94 & 92 & 98 & 91 & 86 & 99 & 98 & 95 & 95 & 93 & 96 & 98 & 98 & 97 \\
 & Prompt-3      & 95 & 92 & 98 & 98 & 93 & 96 & 98 & 92 & 84 & 98 & 93 & 98 & 96 & 90 & 94 & 95 & 90 & 98 \\
 & Multi-hop     & - & - & - & 96 & 85 & 92 & - & - & - & 98 & 87 & 90 & - & - & - & 93 & 96 & 90 \\
\midrule
\multirow{4}{*}{Crime}
 & Prompt-1      & 92 & 85 & 94 & 92 & 87 & 92 & 77 & 84 & 88 & 82 & 77 & 81 & 92 & 89 & 95 & 94 & 96 & 91 \\
 & Prompt-2      & 96 & 95 & 96 & 95 & 98 & 98 & 89 & 92 & 90 & 93 & 90 & 92 & 98 & 95 & 96 & 96 & 98 & 96 \\
 & Prompt-3      & 98 & 92 & 97 & 96 & 96 & 98 & 93 & 94 & 89 & 91 & 91 & 95 & 98 & 96 & 96 & 94 & 93 & 96 \\
 & Multi-hop     & - & - & - & 94 & 88 & 95 & - & - & - & 86 & 81 & 84 & - & - & - & 95 & 96 & 94 \\
\midrule
\multirow{4}{*}{Porn}
 & Prompt-1      & 94 & 85 & 98 & 96 & 91 & 72 & 91 & 88 & 95 & 96 & 80 & 83 & 74 & 81 & 95 & 96 & 96 & 95 \\
 & Prompt-2      & 96 & 93 & 98 & 98 & 98 & 87 & 92 & 93 & 96 & 94 & 98 & 99 & 84 & 88 & 96 & 92 & 97 & 86 \\
 & Prompt-3      & 95 & 95 & 98 & 98 & 93 & 91 & 94 & 93 & 96 & 98 & 97 & 98 & 89 & 87 & 97 & 98 & 95 & 94 \\
 & Multi-hop     & - & - & - & 95 & 96 & 85 & - & - & - & 98 & 87 & 88 & - & - & - & 88 & 93 & 84 \\
\midrule
\multirow{4}{*}{Terrorism}
 & Prompt-1      & 97 & 80 & 99 & 98 & 93 & 83 & 95 & 78 & 98 & 89 & 86 & 78 & 96 & 76 & 98 & 92 & 86 & 81 \\
 & Prompt-2      & 98 & 83 & 99 & 99 & 95 & 90 & 98 & 86 & 99 & 93 & 94 & 83 & 97 & 81 & 99 & 95 & 90 & 86 \\
 & Prompt-3      & 97 & 85 & 96 & 98 & 91 & 92 & 98 & 82 & 98 & 94 & 87 & 81 & 98 & 87 & 99 & 96 & 88 & 89 \\
 & Multi-hop     & - & - & - & 99 & 96 & 89 & - & - & - & 92 & 90 & 85 & - & - & - & 95 & 84 & 87 \\
\midrule
\multirow{4}{*}{Scam}
 & Prompt-1      & 87 & 84 & 96 & 96 & 89 & 84 & 98 & 98 & 99 & 95 & 90 & 89 & 93 & 90 & 97 & 89 & 93 & 86 \\
 & Prompt-2      & 90 & 86 & 93 & 97 & 92 & 85 & 98 & 99 & 98 & 98 & 92 & 93 & 95 & 95 & 98 & 95 & 98 & 91 \\
 & Prompt-3      & 89 & 88 & 95 & 97 & 84 & 87 & 97 & 96 & 98 & 96 & 94 & 90 & 96 & 92 & 91 & 92 & 95 & 89 \\
 & Multi-hop     & - & - & - & 91 & 90 & 85 & - & - & - & 98 & 91 & 94 & - & - & - & 91 & 96 & 92 \\
\midrule
\multirow{4}{*}{Suicide}
 & Prompt-1      & 96 & 86 & 97 & 98 & 90 & 97 & 92 & 90 & 98 & 87 & 87 & 94 & 98 & 88 & 99 & 90 & 89 & 92 \\
 & Prompt-2      & 98 & 89 & 98 & 95 & 94 & 95 & 95 & 93 & 96 & 90 & 93 & 96 & 95 & 89 & 99 & 94 & 90 & 96 \\
 & Prompt-3      & 96 & 88 & 97 & 97 & 91 & 97 & 96 & 91 & 98 & 91 & 88 & 97 & 97 & 93 & 98 & 91 & 86 & 95 \\
 & Multi-hop     & - & - & - & 98 & 92 & 98 & - & - & - & 91 & 89 & 93 & - & - & - & 94 & 90 & 89 \\
\bottomrule
\end{tabular}%
}
\caption{ASR (\%) of \ours on four prompt regimes, namely detection (Prompt-1), description (Prompt-2), summarization (Prompt-3), and multi-hop dialogue. Empty cells mark models without multi-turn support.}
\label{tab:more_prompts_all}
\end{table*}

\begin{table*}[!t]
\centering

\setlength{\tabcolsep}{5pt}
\resizebox{\textwidth}{!}{%
\def\arraystretch{.9}
\begin{tabular}{l|cccccc|cccccc|cccccc}
\toprule
\multirow{2}{*}{VLM}
& \multicolumn{6}{c|}{DKS}
& \multicolumn{6}{c|}{AKS}
& \multicolumn{6}{c}{FRAG} \\
\cmidrule(lr){2-7}\cmidrule(lr){8-13}\cmidrule(lr){14-19}
& Violence & Crime & Porn & Terrorism & Scam & Suicide
& Violence & Crime & Porn & Terrorism & Scam & Suicide
& Violence & Crime & Porn & Terrorism & Scam & Suicide \\
\midrule
BLIP     &  \textbf{93}  &  92  &   94
         &  97  &  87  &  96
         &  \textbf{97}  &  \textbf{77}  &  \textbf{91}
         &  \textbf{95}  &  \textbf{98}  &  \textbf{92}
         &  \textbf{93}  &  \textbf{92}  &  \textbf{74}
         &  \textbf{96}  &  \textbf{93}  &  \textbf{98} \\
CLIP     &  88  &  \textbf{97}  &  94
         &  93  &  82  &  \textbf{97}
         &  86  &  61  &  70
         &  74  &  80  &  67
         &  77  &  81  &  69
         &  83  &  75  &  73   \\
Combined &  91  & 96  &  \textbf{95}
         &  \textbf{98}  &  \textbf{88}  &  93
         &  64  &  70  &  73
         &  79  &  68  &  73
         &  81  &  88  &  72
         &  89  &  80  &  77   \\
\bottomrule
\end{tabular}
}
\caption{ASR (\%) of \ours under three PGS methods when BLIP, CLIP, or their average scores relevance during attack optimization, measured on \LQwen. Bold marks the best surrogate per column.}
\label{tab:ablation_vlm}
\end{table*}

\subsection{More Discussion}\label{sec:more_discussion}

\noindent\textbf{Selection Budget.}
Our target is PGS itself and not the continuously evolving \VideoLLMs that host it. The motivation is that recent \VideoLLMs increasingly replace uniform sampling with PGS-based selection, so models that merely support denser uniform sampling fall outside the scope of this paper. \ours is not tied to a sampling rate either, since the perturbation is applied to every frame of the inserted harmful clip before it replaces a segment of the benign carrier. Even under coarse sampling denser than 1 FPS, PGS must still rank and down-select in order to obtain its top-$N$ prompt-relevant frames, which is exactly the step we attack.

Raising $N$ from 32 to 64 or 128 may intuitively weaken the attack because more harmful frames could survive, and the intuition is wrong for two reasons. Whether a \VideoLLM detects harmful content is governed by the proportion of harmful frames rather than their absolute number, and denser coarse sampling raises benign and harmful counts together without changing that proportion. Many mainstream \VideoLLMs, the LLaVA-based ones in particular, also cannot scale to a much larger $N$ under their own token limits.

We measure this directly by re-running selection at $N \in \{16, 32, 64, 128\}$ on a candidate pool of about 136 frames with everything else fixed. Under AKS the harmful-frame count falls from 1.69 to 0.58 at $N{=}32$ and from 2.86 to 1.31 at $N{=}64$, so the attack behaves the same way across the budgets a deployment would realistically use. At $N{=}128$ the sampler admits 94\% of its own pool, which leaves almost no selection to attack, and the count falls only from 4.00 to 3.25. The other end behaves the same way for the opposite reason, since at $N{=}16$ AKS admits no harmful frames at all with or without the perturbation, so there is nothing for suppression to remove. What governs the attack is therefore the ratio of budget to pool rather than the budget alone, and raising $N$ toward the pool size blunts the attack only by making the sampler stop selecting, which forfeits the very efficiency that had motivated prompt-guided sampling in the first place.

Our evaluation already spans three representative open-source PGS paradigms that differ substantially in relevance estimation and selection, which is what a claim about the family rather than about one implementation requires.

\begin{figure*}[t]
    \centering
    \includegraphics[width=.85\linewidth]{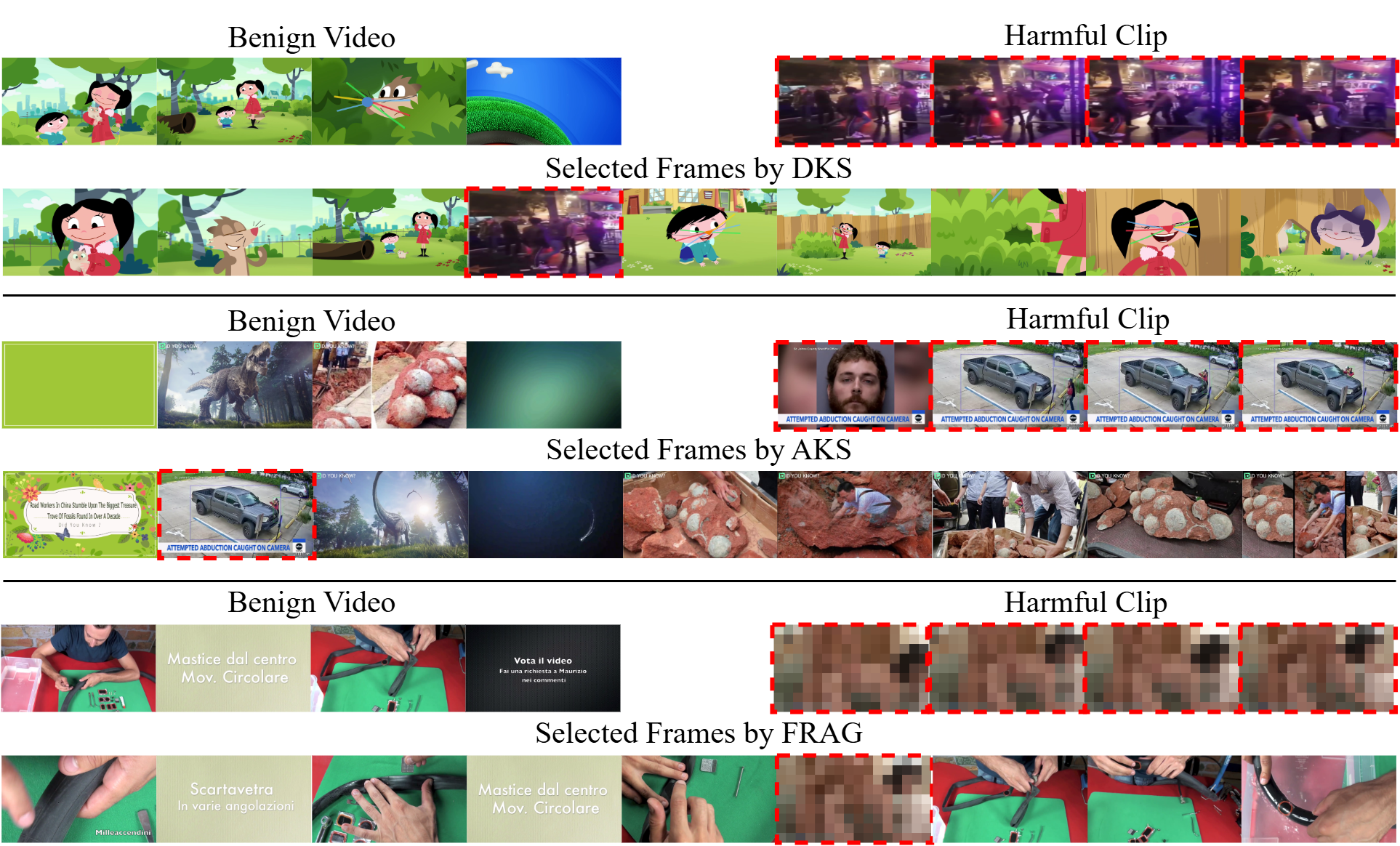}
    \caption{Poisoned videos successfully attacked under \ours, with \textcolor{red}{dashed red boxes} marking harmful frames.}
    \label{fig:examples}
\end{figure*}

\noindent\textbf{Imperceptibility of the Perturbation.}
We constrain the perturbation under $\|\delta\|_\infty \leq 8/255$, the standard threshold in image adversarial attacks at which perturbations are routinely judged imperceptible~\cite{goodfellow2015explaining,madry2018pgd,wong2020fast}. Two video-specific points deserve mention, and the first is that a $\delta$ shared across all frames of the harmful clip rather than resampled per frame introduces no frame-to-frame flicker, so perturbed clips read as a static low-magnitude texture overlay (\Cref{fig:examples}). The threat model also requires only that the harmful event stay recognizable to a human who watches the clip, since the attack's leverage comes from keeping those frames out of the model input, so disguising them is unnecessary and would in fact work against us.

\Cref{tab:imperceptibility} reports the resulting distortion on the harmful clips. Absolute values of SSIM near 0.84 and LPIPS near 0.31 may look far from the clean clip for a budget described as imperceptible, so \Cref{tab:quality_calibration} places every attack we study on the same axis at the same budget. Feature collapse there denotes a text-free attack that pushes the frame embedding away from its clean position, the generic way to attack a vision-language encoder. Random Gaussian noise, which leaves attack success at the level of plain frame replacement, already costs 0.909 SSIM and 0.220 LPIPS, and one-step FGSM costs 0.793 and 0.369, while \ours sits between them at 0.837 and 0.314. Most of the distortion these metrics report therefore comes from filling an $\ell_\infty$ ball of this radius on dense video, and the part attributable to optimizing rather than sampling the perturbation is the gap to the Gaussian row. We report this openly because it bounds how far the imperceptibility claim reaches. No perceptual experiment on the perturbation itself was run for this paper. The figures above are objective full-reference metrics, and the statement that the depicted event stays identifiable rests on inspection by the authors rather than on ratings collected from recruited observers, which is a weaker basis and we do not present it as more. A study with recruited observers would sharpen the claim and we leave it to future work, and it does not carry the argument here, since the failure we expose is omission at the moderator output and human-side imperceptibility plays no part in it. Retaining human recognizability is a \emph{stricter} requirement in our threat model than allowing the harmful content to be obscured by the perturbation.

\begin{table}[t]
\centering

\setlength{\tabcolsep}{5pt}
\resizebox{\columnwidth}{!}{%
\def\arraystretch{.95}
\begin{tabular}{lcccc}
\toprule
\textbf{Attack} & \textbf{SSIM} $\uparrow$ & \textbf{PSNR} $\uparrow$ & \textbf{LPIPS} $\downarrow$ & \textbf{Effective on PGS} \\
\midrule
Random Gaussian  & 0.909 & 33.1 & 0.220 & no \\
Feature collapse & 0.856 & 32.8 & 0.297 & partly \\
\ours            & 0.837 & 32.6 & 0.314 & yes \\
FGSM             & 0.793 & 30.3 & 0.369 & no \\
\bottomrule
\end{tabular}
}
\caption{Distortion introduced by each attack under a matched perturbation budget on the harmful clips.}
\label{tab:quality_calibration}
\end{table}

\begin{table}[t]
\centering

\setlength{\tabcolsep}{4pt}
\resizebox{\columnwidth}{!}{%
\def\arraystretch{.95}
\begin{tabular}{lcccccc}
\toprule
\textbf{Type} & \textbf{SSIM} & \textbf{LPIPS} & \textbf{DISTS} & \textbf{KID Mean} & \textbf{KID Std} & \textbf{FID} \\
\midrule
Violence  & 0.82 & 0.30 & 0.10 & 0.04 & 0.01 & 76.10 \\
Crime     & 0.79 & 0.39 & 0.12 & 0.06 & 0.01 & 72.02 \\
Porn      & 0.82 & 0.30 & 0.11 & 0.12 & 0.00 & 91.37 \\
Terrorism & 0.78 & 0.43 & 0.17 & 0.11 & 0.00 & 109.77 \\
Scam      & 0.82 & 0.28 & 0.11 & 0.06 & 0.00 & 80.63 \\
Suicide   & 0.82 & 0.22 & 0.10 & 0.09 & 0.00 & 54.66 \\
\bottomrule
\end{tabular}
}
\caption{Imperceptibility between original and perturbed harmful videos. Higher SSIM indicates better visual similarity, while lower LPIPS, DISTS, KID, and FID indicate smaller perceptual distortion, averaged within category.}
\label{tab:imperceptibility}
\end{table}

\noindent\textbf{Closed-source and Proprietary VideoLLMs.}
Our experiments cover six representative open-source \VideoLLMs, and closed-source systems also exist~\cite{hurst2024gpt4o,team2023gemini,team2024reka}. Their technical reports describe uniform sampling with no prompt-guided selection~\cite{team2023gemini}, and the sampling stage is exposed neither for inspection nor for replacement, so neither the attack surface we study nor the measurement we need is available in them. Direct evaluation on PGS-based commercial systems is therefore not feasible at present. The goal of this work is nonetheless to flag the safety risk inherent in PGS in a timely way, as an early reminder for developers weighing this increasingly popular sampler for future \VideoLLMs.

\noindent\textbf{Number of Benign Videos.}
We evaluate on 200 benign carriers, which, given six harmful categories and the randomized temporal position of each insertion, is already a substantial sample for a video-based evaluation. During the early proof-of-concept stage we also ran 1{,}000 randomly selected benign videos when attacking \LQwen under the three PGS methods, and the results agreed with those from 200. Since the goal is to expose a safety implication inherent in current PGS methods, the number of benign carriers does not materially influence any of the conclusions that we draw from them.

\noindent\textbf{Long Video Understanding.}
The community has no sharp boundary in duration between long and short video understanding. Papers describing their models as long-video systems~\cite{tang2025aks,huang2025frag,weng2024longvlm,cheng2024focuschat} typically evaluate on videos of several tens of seconds to one or two minutes, whereas short-video understanding usually involves a few seconds, and some recent work~\cite{qian2024streaming,wang2025videotree} claims tens of minutes to an hour. The PGS methods considered here are built for one or two minutes up to several tens of minutes, which matches the common definition of long-video understanding, and longer videos demand more precise sampling to locate the scenes a prompt refers to. Our experiments show that \ours keeps harmful frames out of the selection on exactly these models. Hour-long video is an interesting direction for future work, and since longer carriers admit longer harmful clips, we expect the attack there to grow more consequential rather than less as videos lengthen.

\noindent\textbf{Impact of Video Length.}
Carrier length affects attack performance through two distinct mechanisms. With the harmful clip fixed, a longer carrier lowers the proportion of harmful content and so the probability that harmful frames are sampled at all, and the weakness there is statistical, since harmful content becomes harder to sample regardless of how sophisticated the sampler is. When the harmful clip may grow instead, an hour-long carrier admits a minute-long harmful clip, which preserves the applicability of \ours while amplifying the visibility and the impact of the harmful content for human observers. Across both regimes, the \VideoLLMs we study show a limited ability to surface harmful signals reliably, and in both regimes the sampler is the reason why.

\subsection{Limitations and Open Questions}
The main text states the setting in which our claims hold, and this section records what that setting leaves unresolved.

\noindent\textbf{Reach of the Adversary.}
\ours assumes the attacker can perturb the pixels it uploads. Adversaries confined to metadata, to the audio track, or to sub-frame regions are not covered, and whether relevance can be suppressed under those narrower controls is open.

\noindent\textbf{Coverage of the Depiction Set.}
Our depiction sets come from one shadow \VideoLLM and one paraphrasing language model. The validation above finds the captioning stage to be the weaker of the two, since four of 60 base descriptions name the wrong act, and harmful semantics that either model describes poorly should yield weaker depictions still. How far that reaches is not something our six categories can measure.

\noindent\textbf{Video Length.}
We study minute-scale video, which matches the deployments PGS was built for. Hour-scale and live-streaming inputs raise questions of their own, since a sampler that streams cannot rank a pool it has not yet seen.

\noindent\textbf{Adaptive Defenses.}
\Cref{tab:defenses} evaluates seven non-adaptive defenses, so the residual robustness it reports is an upper bound. A proper adaptive comparison would give each sampling-stage defense its own study, which is work we leave to follow-up rather than sketch here.

\noindent\textbf{Scale of the Selection Measurements.}
The frame-level measurements in this appendix come from our own reimplementation of the samplers at reduced scale, and the relevance scorer we use for FRAG need not match the one behind \Cref{tab:main_results}. How much of FRAG's attack success is due to eviction rather than to the downstream model is therefore not something these numbers alone can settle.

\noindent\textbf{Ingestion.}
\Cref{tab:ingestion} shows that downscaling on ingest gives back part of what the perturbation removed and still leaves ASR at 46. An adversary facing a real transcoding ladder would have to optimize through it, which we have not done.

\subsection{Ethical Considerations}
This work examines a safety vulnerability in \VideoLLMs deployed as content moderators, and we treat it as high-risk dual-use research. This section records where the harmful material came from and how it was handled, what an adversary gains from the paper, and what we release, to whom, on what conditions, and after how long a delay.

\noindent\textbf{Provenance of the Harmful Clips.}
The evaluation uses 60 harmful clips, 10 in each of six categories, drawn from two kinds of source that we handle differently. One part comes from public research datasets, namely RLVS~\cite{soliman2019violence} and XD-Violence~\cite{wu2020xd-violence}, which we use under the research terms their authors state and cite in full. The rest are short segments of footage posted publicly on online video platforms and retrieved by URL. That second group carries no license permitting redistribution, and no consent was or could be obtained from the people depicted, since the recordings are third-party uploads that we neither commissioned nor control. We therefore treat those clips as non-redistributable evaluation inputs. They serve only as model inputs and never as training data, they are held on access-controlled storage available to the authors alone, and our records identify them by source and timestamp rather than copying them into any released artifact. The cost to reproducibility is real and we state it rather than paper over it, in that the exact clip set cannot be rebuilt from what we release, and we judge it correct for material of this sort.

\noindent\textbf{Human Participants.}
No participants were recruited from outside the author group. The depiction sets were validated by three of the authors rating independently, as reported above, and clip selection, screening, and the visual inspection under imperceptibility were carried out by the authors as well, who are the only people who viewed the harmful material. We employed no annotators and no crowdworkers and collected no ratings or personal data from third parties, which is a deliberate choice for material of this kind rather than a shortcut, since showing it to recruited workers would transfer onto them the harm we study. No perceptual experiment on the perturbation was run, which is why every imperceptibility number is an objective metric.

\noindent\textbf{Misuse Analysis.}
We state the capability the paper confers precisely enough to be weighed. Against a moderator built on an open PGS implementation, an uploader holding a single commodity GPU can suppress a four-second harmful insert at between 84\% and 97\% success, at under two minutes of offline optimization per clip and with no access to the moderator at all. Three properties bound it, the first being that the attack confers nothing on a pipeline that reads every frame, because there is no ranking to poison. A pipeline whose ingest resamples the upload takes 48 points off it, which \Cref{tab:ingestion} measures, and that is a reduction rather than a fix. On uniform and semantic samplers the omission it produces is the one \citet{cao2026failures} already documented, so the perturbation adds little there that frame replacement did not achieve alone. Exposure therefore concentrates on platforms that have adopted prompt-guided selection, and resampling on ingest narrows that population without emptying it, which is why our disclosure targets it rather than the public.

\noindent\textbf{Release and Gating.}
The release is tiered rather than open, because an ungated artifact is not proportionate to the capability just described. Without restriction we publish the evaluation protocol, the benign carrier identifiers, the prompt templates, and the analysis scripts, which together suffice to check every number in this paper. The relevance-suppression optimization code is released only after the disclosure window below has closed, and then through a request process that records the requester's identity and institution and requires a written statement of research use. We release at no point the harmful clips, the perturbations optimized against them, the depiction sets, or the per-clip success traces, since those together form a working evasion kit and none is needed to verify a claim we make. Gating slows independent replication, and we accept that as the smaller cost.

\noindent\textbf{Coordinated Disclosure.}
At the time of submission no disclosure had been made, because the paper is under anonymous review and contacting maintainers would identify us. We state this rather than imply that notification has already happened. On acceptance we will notify the maintainers of the three open PGS implementations we evaluate, together with the vendors of the six \VideoLLMs we test, and we will hold the gated code for ninety days from that notification before opening the request process. Each notification will carry the sampling-stage defenses developed here and the practitioner measures given in the main text, so that a recipient receives a mitigation with the report rather than after it.

\noindent\textbf{On Publishing the Adaptive Attack.}
The main text observes that an adaptive attacker could fold the auxiliary scorers of ERS, or the classifier of VLM-Det, into the relevance-suppression objective, and a reader may fairly ask whether that sentence should have been written. We think it should, because those two defenses are the only ones that recover more than a few points, at 15.7 and 8.0, and a reader who saw the non-adaptive numbers alone could deploy either believing it a fix. Naming the limit is what prevents that outcome, and we give a direction rather than an implementation, since we neither built nor evaluated the adaptive attack and no adaptive result appears anywhere in this paper, so what the passage transfers is a consequence of the threat model rather than working code.

\noindent\textbf{Why We Publish.}
PGS is being adopted quickly in production \VideoLLMs while its safety surface stays unstudied. Withholding the finding would leave deployers unaware of a weakness that an independent party can rediscover, and it would delay the defenses that become visible only once the attack is written down. We judge disclosure, paired with the gating described above and with the concrete mitigations given in the main text, to be by far the better outcome here.

\end{document}

%% file: math_commands.tex
\usepackage{amsmath,amsfonts,bm}

\def\eqref#1{equation~\ref{#1}}

\def\1{\bm{1}}

\DeclareMathAlphabet{\mathsfit}{\encodingdefault}{\sfdefault}{m}{sl}
\SetMathAlphabet{\mathsfit}{bold}{\encodingdefault}{\sfdefault}{bx}{n}

%% file: _header.tex
\usepackage{graphicx}
\usepackage[table]{xcolor}
\usepackage{tikz}
\usepackage{amsfonts}
\usepackage{xspace} 
\usepackage{amsmath}
\usepackage{mathrsfs}
\usepackage{amssymb}
\usepackage{enumitem}
\usepackage{multirow}
\usepackage{makecell}
\usepackage{caption}
\usepackage{subcaption}
\usepackage{booktabs}
\usepackage{tcolorbox}
\usepackage{xurl}
\usepackage{tabularx}

\usepackage{diagbox}
\usepackage{placeins}
\setlist[itemize]{leftmargin=3mm}
\usepackage{dsfont}
\usepackage{cleveref}



\usepackage{siunitx}
\def\ie{\emph{i.e.,}\xspace}